%% file: main.tex
\documentclass[10pt,twocolumn,letterpaper]{article}

\usepackage{cvpr}
\usepackage{times}
\usepackage{epsfig}
\usepackage{graphicx}
\usepackage{amsmath}
\usepackage{amssymb}
\usepackage{svg}
\usepackage{float}

\usepackage[font=small,labelfont=bf,tableposition=top]{caption} % for the teaser

\usepackage[pagebackref=true,breaklinks=true,letterpaper=true,colorlinks,bookmarks=false]{hyperref}

\cvprfinalcopy % *** Uncomment this line for the final submission

\makeatletter
\let\@authortofile\@empty
\let\@affiltofile\@empty
\def\@separator{\def\@separator{,\hspace{6pt}}}% Delay comma once
\def\addauthor#1#2{%
	\g@addto@macro\@author{\@separator#1\textsuperscript{#2}}
	\g@addto@macro\@authortofile{\@separator#1}
}
\def\affilnocomma#1#2{%
	\g@addto@macro\@author{\\[0.2cm] \@separator$^#2$#1}
	\g@addto@macro\@affiltofile{\@separator#1}
}
\def\affil#1#2{%
	\g@addto@macro\@author{\@separator$^#2$#1}
	\g@addto@macro\@affiltofile{\@separator#1}
}
\def\url#1{%
	\g@addto@macro\@author{\\[0.1cm]#1}
	\g@addto@macro\@affiltofile{#1}
}
\def\blfootnote{\gdef\@thefnmark{}\@footnotetext}
\makeatother

\graphicspath{{figures/}}

\ifcvprfinal\fi
\begin{document}

%%%%%%%%% TITLE
\title{DeepVoxels: Learning Persistent 3D Feature Embeddings}

\addauthor{Vincent Sitzmann}{1}
\addauthor{Justus Thies}{2}
\addauthor{Felix Heide}{3}
\addauthor{\vspace{0.07cm}\\}{}
\addauthor{Matthias Nie{\ss}ner}{2}
\addauthor{Gordon Wetzstein}{1}
\addauthor{Michael Zollh\"ofer}{1}
\affilnocomma{Stanford University}{1}
\affil{Technical University of Munich}{2}
\affil{Princeton University}{3}
\url{\tt\normalsize \href{https://vsitzmann.github.io/deepvoxels/}{vsitzmann.github.io/deepvoxels/}}

\maketitle

\pagenumbering{gobble}

\begin{abstract}

In this work, we address the lack of 3D understanding of generative neural networks by introducing a persistent 3D feature embedding for view synthesis.
To this end, we propose \textit{DeepVoxels}, a learned representation that encodes the view-dependent appearance of a 3D scene without having to explicitly model its geometry.
At its core, our approach is based on a Cartesian 3D grid of persistent embedded features that learn to make use of the underlying 3D scene structure.
Our approach combines insights from 3D geometric computer vision with recent advances in learning image-to-image mappings based on adversarial loss functions.
DeepVoxels is supervised, without requiring a 3D reconstruction of the scene, using a 2D re-rendering loss and enforces perspective and multi-view geometry in a principled manner.
We apply our persistent 3D scene representation to the problem of novel view synthesis demonstrating high-quality results for a variety of challenging scenes.
\end{abstract}
\vspace{-0.6cm}

\section{Introduction}
\label{sec:intro}
\input{sections/introduction}
\section{Related Work}
\label{sec:related}
\input{sections/related}

\section{Method}
\label{sec:method}
\input{sections/overview}

\section{Analysis}
\label{sec:analysis}
\input{sections/analysis}

\section{Limitations}
\label{sec:limitations}
\input{sections/limitations}

\section{Conclusion}
\input{sections/discussion}

\input{sections/acknowledgements}

{\small
\bibliographystyle{ieee}
\bibliography{references}
}

\end{document}

%% file: sections/introduction.tex
Recent years have seen significant progress in applying generative machine learning methods to the creation of synthetic imagery.
Many deep neural networks, for example based on (variational) autoencoders, are able to inpaint, refine, or even generate complete images from scratch \cite{HintoS2006,jKingma2014}.
A very prominent direction is generative adversarial networks \cite{GoodfPMXWOCB2014} which achieve impressive results for image generation, even at high resolutions \cite{KarraALL2018} or conditional generative tasks \cite{IsolaZZE2017}.
These developments allow us to perform highly-realistic image synthesis in a variety of settings; e.g., purely generative, conditional, etc.

However, while each generated image is of high quality, a major challenge is to generate a series of coherent views of the same scene.
Such consistent view generation would require the network to have a latent space representation that fundamentally understands the 3D layout of the scene; e.g., how would the same chair look from a different viewpoint?
Unfortunately, this is challenging to learn for existing generative neural network architectures that are based on a series of 2D convolution kernels.
Here, spatial layout and transformations of a real, 3D environment would require a tedious learning process which maps 3D operations into 2D convolution kernels \cite{Jaderberg2015}.
In addition, the generator network in these approaches is commonly based on a U-Net architecture with skip connections \cite{RonneFB2015}.
Although skip connections enable efficient propagation of low-level features, the learned 2D-to-2D mappings typically struggle to generalize to large 3D transformations, due to the fact that the skip connections bypass higher-level reasoning.

\begin{figure}[t]
	\centering
	\includegraphics[width=\linewidth]{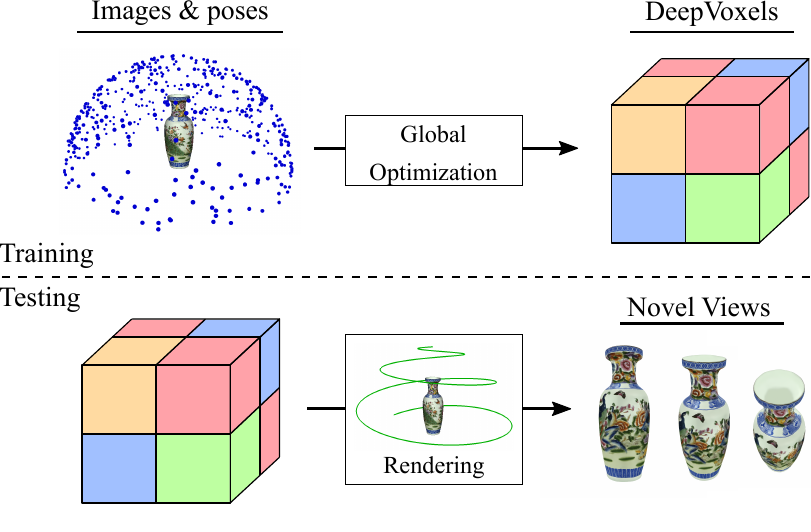}
	\vspace{-0.8cm}
	\caption
	{
		During training, we learn a persistent DeepVoxels representation that encodes the view-dependent appearance of a 3D scene from a dataset of posed multi-view images (top).
		At test time, DeepVoxels enable novel view synthesis (bottom).
	}
	\label{fig:corpus}
	\vspace{-0.3cm}
\end{figure}

To tackle similar challenges in the context of learning-based 3D reconstruction and semantic scene understanding, the field of 3D deep learning has seen large and rapid progress over the last few years.
Existing approaches are able to predict surface geometry with high accuracy.
Many of these techniques are based on explicit 3D representations in the form of occupancy grids \cite{Maturana2015,qi2016volumetric}, signed distance fields \cite{Riegler2017OctNet}, point clouds \cite{qi2017pointnet, lin2018learning}, or meshes \cite{Jack2018}.
While these approaches handle the geometric reconstruction task well, they are not directly applicable to the synthesis of realistic imagery, since it is unclear how to represent color information at a sufficiently high resolution.
There also exists a large body of work on learning low-dimensional embeddings of images that can be decoded to novel views~\cite{tatarchenko2015single,yan2016perspective,dosovitskiy2017learning,eslami2018neural,worrall2017interpretable,rhodin2018unsupervised}.
Some of these techniques make use of the object's 3D rotation by explicitly rotating the latent space feature vector \cite{worrall2017interpretable,rhodin2018unsupervised}.
While such 3D techniques are promising, they have thus far not been successful in achieving sufficiently high fidelity for the task of photorealistic image synthesis.

In our work, we aim at overcoming the fundamental limitations of existing 2D generative models by introducing native 3D operations in the neural network architecture.
Rather than learning intuitive concepts from 3D vision, such as perspective, we explicitly encode these operations in the network architecture and perform reasoning directly in 3D space.
The goal of the DeepVoxels approach is to condense posed input images of a scene into a persistent latent representation without explicitly having to model its geometry (see Fig.~\ref{fig:corpus}).
This representation can then be applied to the task of novel view synthesis to generate unseen perspectives of a 3D scene without requiring access to the initial set of input images.
Our approach is a hybrid 2D/3D one in that it learns to represent a scene in a Cartesian 3D grid of persistent feature embeddings that is projected to the target view's canonical view volume and processed by a 2D rendering network.
This persistent feature volume, which exists in 3D world-space, in combination with a structured, differentiable image formation model, enforces perspective and multi-view geometry in a principled and interpretable manner during training.
The proposed approach learns to exploit the underlying 3D scene structure, without requiring supervision in the 3D domain.
We demonstrate novel view synthesis with high quality for a variety of scenes based on this new representation.
In summary, our approach makes the following technical contributions:
\begin{itemize}
    \itemsep 0em 
	\item A novel persistent 3D feature representation for image synthesis that makes use of the underlying 3D scene information.
	\item Explicit occlusion reasoning based on learned soft visibility that leads to higher-quality results and better generalization to novel viewpoints.
	\item Differentiable image formation to enforce perspective and multi-view geometry in a principled and interpretable manner during training.
	\item Training without requiring 3D supervision.
\end{itemize}

\paragraph{Scope}
In this paper, we present first steps towards 3D-structured neural scene representations.
To this end, we limit the scope of our investigation to allow an in-depth discussion of the challenges fundamental to this approach.
We assume Lambertian scenes, without specular highlights or other view-dependent effects. 
While the proposed approach can deal with light specularities, these are not modeled explicitly. 
Classical approaches will achieve impressive results on the presented scenes. However, these approaches rely on the explicit reconstruction of geometry. 
Neural scene representations will be essential to develop generative models that can generalize across scenes to solve reconstruction problems where only few observations are available. 
We thus compare to such baselines exclusively.

%% file: sections/related.tex
Our approach lies at the intersection of multiple active research areas, namely generative neural networks, 3D deep learning, deep learning-based view synthesis, and model- as well as image-based rendering.

\paragraph{Neural Image Synthesis}
Deep models for 2D image and video synthesis have recently shown very promising results.
Some of these approaches are based on (variational) auto-encoders (VAEs) \cite{HintoS2006,jKingma2014} or autoregressive models (AMs), such as PixelCNN \cite{Oord:2016}.
The most promising results so far are based on conditional generative adversarial networks (cGANs) \cite{GoodfPMXWOCB2014,RadfoMC2016,MirzaO2014,IsolaZZE2017}.
In most cases, the generator network has an encoder-decoder architecture \cite{HintoS2006}, often with skip connections (U-Net) \cite{RonneFB2015}, which enable efficient propagation of low-level features from the encoder to the decoder.
Approaches that convert synthetic images into photo-realistic imagery have been proposed for the special case of human bodies \cite{Zhu2018,Chan2018} and faces \cite{kim2018DeepVideo}.
In theory, similar architectures could be used to regress the real-world image corresponding to a given viewpoint, i.e., image-based rendering could be learned from scratch.
Unfortunately, these 2D-to-2D translation approaches struggle to generalize to transformations in 3D space, such as rotation and perspective projection, since the underlying 3D scene structure cannot be exploited.
We compare to this baseline in Sec.~\ref{sec:analysis} and show that DeepVoxels drastically outperforms it.

\paragraph{3D Deep Learning}
Recently, deep learning has been successfully applied to many 3D geometric reasoning tasks.
Current approaches are able to predict an accurate 3D representation of an object from just a single or multiple views.
Many of these techniques make use of classical 3D representations, e.g., occupancy grids \cite{Maturana2015,qi2016volumetric}, signed distance fields \cite{Riegler2017OctNet}, 3D point clouds \cite{qi2017pointnet, lin2018learning}, or meshes \cite{Jack2018}.
While these approaches handle the geometric reconstruction task well, they are not directly applicable to view synthesis, since it is unclear how to represent color information at a sufficiently high resolution.
View consistency can be explicitly handled using differentiable ray casting \cite{drcTulsiani17}.
RenderNet~\cite{RenderNet} learns to render in different styles from 3D voxel grid input.
Kulkarni et al.~\cite{Kulkarni2015} learn a disentangled representation of images with respect to various scene properties, such as rotation and illumination.
Spatial Transformer Networks \cite{Jaderberg2015} can learn spatial transformations of feature maps in the network.
Even weakly-supervised \cite{yang2015weakly} and unsupervised \cite{NIPS2016_6600} learning of 3D transformations has been proposed.
Our work is also related to CNNs for 3D reconstruction~\cite{kar2017learning, choy20163d} and monocular depth estimation~\cite{eigen2014depth}.
A ``multi-view stereo machine''~\cite{kar2017learning} can learn 3D reconstruction based on 3D or 2.5D supervision.
MapNet~\cite{henriques2018mapnet} performs SLAM based on a scene-specific 2D feature grid representation.
In contrast to these approaches, which are focused on geometric reasoning, our goal is to learn an embedding for novel view synthesis. To synthesize multi-view consistent images, we optimize for a persistent, scene-specific 3D embedding over all available 2D observations and enable the network to perform explicit occlusion reasoning. We do not require any 3D ground truth but minimize a 2D photometric reprojection loss exclusively.

\paragraph{Deep Learning for View Synthesis}
Recently, a class of deep neural networks has been proposed that directly aim to solve the problem of novel view synthesis.
Some techniques predict lookup tables into a set of reference views~\cite{Park17,zhou2016view} or predict weights to blend multi-view images into novel views~\cite{flynn2016deepstereo}.
A layered scene representation \cite{lsiTulsiani18} can be learned based on a re-rendering loss.
A large corpus of work focuses on embedding 2D views of scenes into a learned low-dimensional latent space that is then decoded into a novel view~\cite{tatarchenko2015single,yan2016perspective,dosovitskiy2017learning,eslami2018neural,worrall2017interpretable,rhodin2018unsupervised,cohen2014transformation}. 
Some of these approaches rely on embedding views into a latent space that does not enforce any geometrical constraints~\cite{tatarchenko2015single,dosovitskiy2017learning, eslami2018neural}, others enforce geometric constraints in varying degrees~\cite{worrall2017interpretable,rhodin2018unsupervised,cohen2014transformation,falorsi2018explorations}, such as learning rotation-equivariant features by explicitly rotating the latent space feature vectors.
We focus on optimizing a scene-specific embedding over a training corpus of 2D observations and explicitly account for concepts from 3D vision such as perspective projection and occlusion to constrain the latent space. We demonstrate advantages over weakly structured embeddings in generating high-quality novel views.

\paragraph{Model-Based Rendering}
Classic reconstruction approaches such as structure-from-motion exploit multi-view geometry \cite{Hartley:2003, szeliski2010computer} to build a dense 3D point cloud of the imaged scene~\cite{schoenberger2016sfm, schoenberger2016mvs, snavely2006photo, agarwal2009building, furukawa2010accurate}.
A triangular surface representation can be obtained using for example the Poisson Surface \cite{Kazhdan:2006} reconstruction technique.
However, the reconstructed geometry is often imperfect, coarse, contains holes, and the resulting renderings thus suffer from visible artifacts and are not fully realistic.
In contrast, our goal is to learn a representation that efficiently encodes the view-dependent appearance of a 3D scene without having to explicitly reconstruct a geometric model.

\paragraph{Image-Based Rendering}  
Traditional image-based rendering techniques blend warped versions of the input images to generate new views~\cite{shum2000review}.
This idea was first proposed as a computationally efficient alternative to classical rendering~\cite{lippman1980movie, greene1986environment, chen1993view}.
Multiple-view geometry can be used to obtain the geometry for warping~\cite{hedman2016scalable}.
In other cases, no 3D reconstruction is necessary~\cite{flynn2016deepstereo, penner2017soft}.
Some approaches rely on light fields~\cite{kalantari2016learning}. 
Recently, deep-learning has been used to aid image-based rendering via learning a small subtask, i.e., the computation of the blending weights~\cite{hedman2018deepblending, flynn2016deepstereo}.
While this can achieve photorealism, it depends on a dense set of high-resolution photographs to be available at rendering time and requires an error prone reconstruction step to obtain the geometric proxy.
Our approach has orthogonal goals: (1) we want to learn an embedding for view synthesis and (2) we want to tackle the problem in a holistic fashion by learning raw pixel output.
Thus, our approach is more related to embedding techniques that try to learn a latent space that can be decoded into novel views.

%% file: sections/overview.tex
\begin{figure*}[t!]
	\centering
	\includegraphics[width=\textwidth]{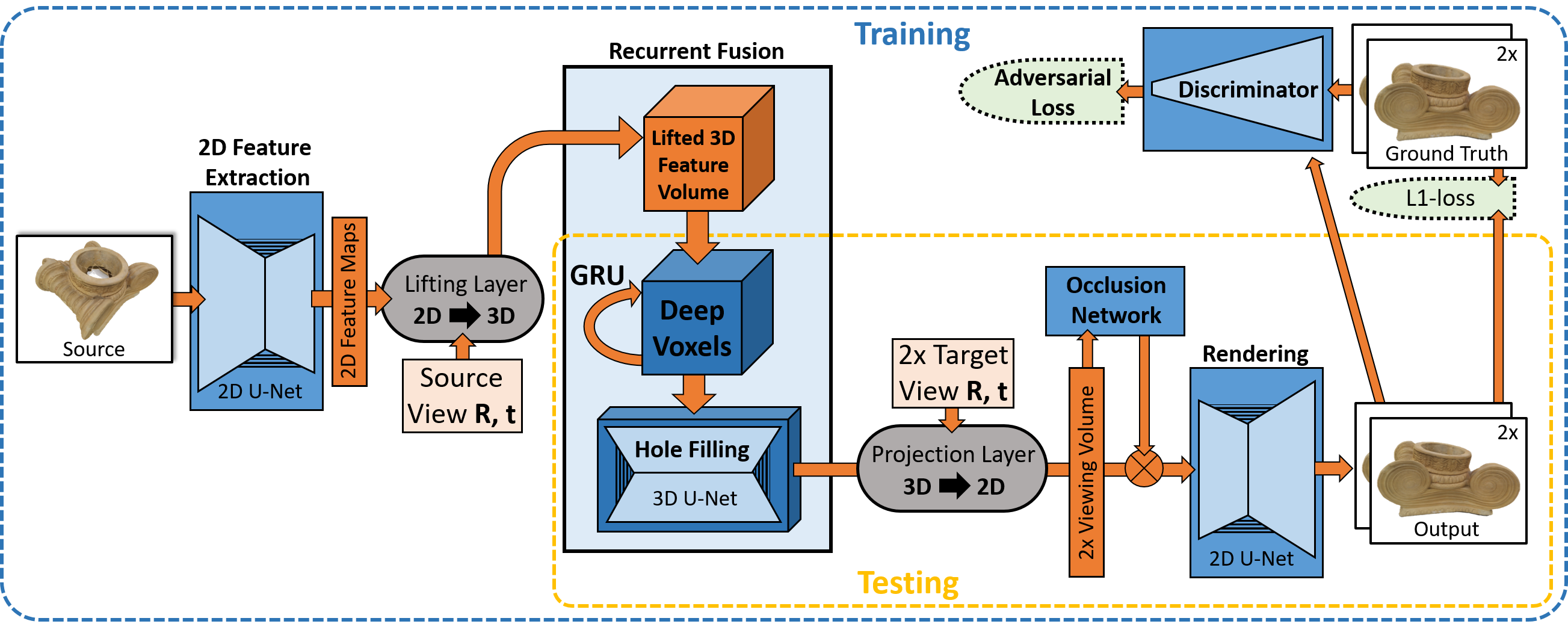}
	\vspace{-0.6cm}
	\caption{Overview of all model components. At the heart of our encoder-decoder based architecture is a novel viewpoint-invariant and persistent 3D volumetric scene representation called DeepVoxels that enforces spatial structure on the learned per-voxel code vectors.}
	\vspace{-0.3cm}
	\label{fig:pipeline_overview}
\end{figure*}

The core of our approach is a novel 3D-structured scene representation called DeepVoxels. 
DeepVoxels is a viewpoint-invariant, persistent and uniform 3D voxel grid of features.
The underlying 3D grid enforces spatial structure on the learned per-voxel code vectors.
The final output image is formed based on a 2D network that receives the perspective re-sampled version of this 3D volume, i.e., the canonical view volume of the target view, as input.
The 3D part of our approach takes care of spatial reasoning, while the 2D part enables fine-scale feature synthesis.
In the following, we first introduce the training corpus and then present our end-to-end approach for finding the scene-specific DeepVoxels representation from a set of multi-view images without explicit 3D supervision.

\subsection{Training Corpus}

Our scene-specific training corpus $\mathcal{C} = \{\mathcal{S}_i, \mathcal{T}_i^0, \mathcal{T}_i^1 \}_{i=1}^{M}$ of $M$ samples is based on a source view $\mathcal{S}_i$ (image and camera pose) and two target views $\mathcal{T}_i^0,\mathcal{T}_i^1$, which are randomly selected from a set of $N$ registered multi-view images; see Fig.~\ref{fig:corpus} for an example.
We assume that the intrinsic and extrinsic camera parameters are available.
These can for example be obtained using sparse bundle adjustment \cite{Triggs:1999}.
For each pair of target views $\mathcal{T}_i^0,\mathcal{T}_i^1$ we then randomly select a single source view $\mathcal{S}_i$ from the top-$5$ nearest neighbors in terms of view direction angle to target view $\mathcal{T}_i^0$.
This sampling heuristic makes it highly likely that points in the source view are visible in the target view $\mathcal{T}_i^0$. While not essential to training, this ensures meaningful gradient flow for every optimization step, while encouraging multi-view consistency to the random target view $\mathcal{T}_i^1$.
We sample the training corpus $\mathcal{C}$ dynamically during training.

\subsection{Architecture Overview}
\begin{figure*}[t]
	\centering
	\includegraphics[width=\linewidth]{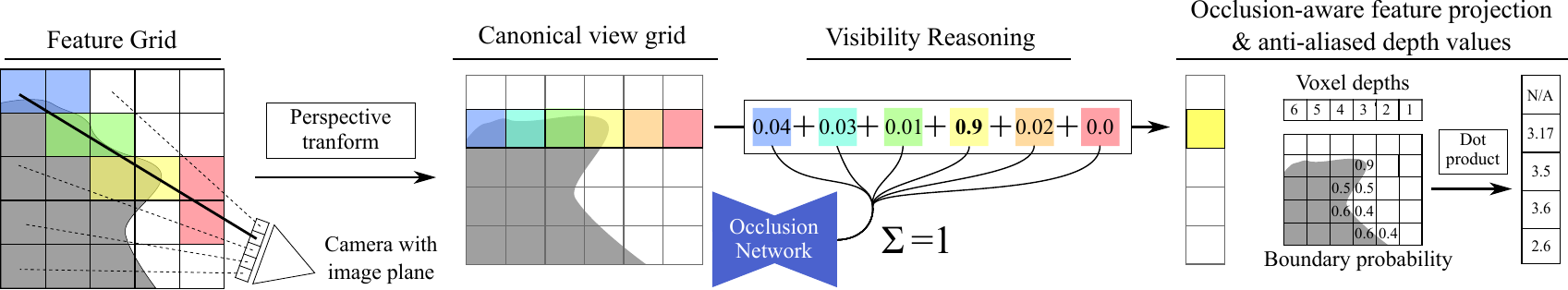}
	\caption{
		Illustration of the occlusion-aware projection operation.
		The feature volume (represented by feature grid) is first resampled into the canonical view volume via a projection transformation and trilinear interpolation.
		The occlusion network then predicts per-pixel softmax weights along each depth ray.
		The canonical view volume is then collapsed along the depth dimension via a softmax-weighted sum of voxels to yield the final, occlusion-aware feature map.
		The per-voxel visibility weights can be used to compute a depth map.
	}
	\label{fig:occlusion_net}
\end{figure*}

Our network architecture is summarized in Fig.~\ref{fig:pipeline_overview}. 
On a high level, it can be seen as an encoder-decoder based architecture with the persistent 3D DeepVoxels representation as its latent space.
During training, we feed a source view $\mathcal{S}_i$ to the encoder and try to predict the target view $\mathcal{T}_i$.
We first extract a set of 2D feature maps from the source view using a 2D feature extraction network.
To learn a view-independent 3D feature representation, we explicitly lift image features to 3D based on a differentiable lifting layer.
The lifted 3D feature volume is fused with our persistent DeepVoxels scene representation using a gated recurrent network architecture.
Specifically, the persistent 3D feature volume is the hidden state of a gated recurrent unit (GRU)~\cite{Cho2014}.
After feature fusion, the volume is processed by a 3D fully convolutional network.
The volume is then mapped to the camera coordinate systems of the two target views via a differentiable reprojection layer, resulting in the canonical view volume.
A dedicated, structured occlusion network operates on the canonical view volume to reason about voxel visibility and flattens the view volume to a 2D view feature map (see Fig.~\ref{fig:occlusion_net}).
Finally, a learned 2D rendering network forms the two final output images.
Our network is trained end-to-end, without the need of supervision in the 3D domain, by a 2D re-rendering loss that enforces that the predictions match the target views.
In the following, we provide more details.

\paragraph{Camera Model}
We follow a perspective pinhole camera model that is fully specified by its extrinsic $\mathbf{E} = \big[\mathbf{R}|\mathbf{t}\big] \in \mathbb{R}^{3\times4}$ and intrinsic $\mathbf{K} \in \mathbb{R}^{3\times3}$ camera matrices \cite{Hartley:2003}.
Here, $\mathbf{R} \in \mathbb{R}^{3\times3}$ is the global camera rotation and $\mathbf{t} \in \mathbb{R}^{3}$ its translation.
Assume we are given a position $\mathbf{x} \in \mathbb{R}^3$ in 3D coordinates, then the mapping from world space to the canonical camera volume is given as:
\begin{equation}
\label{eq:projection}
\mathbf{u}
=
\left(
\begin{array}{c}
u\\
v\\
d\\
\end{array}
\right)
=
\mathbf{K}(\mathbf{R}\mathbf{x}+\mathbf{t})
\enspace{.}
\end{equation}
Here, $u$ and $v$ specify the position of the voxel center on the screen and $d$ is its depth from the camera.
Given a pixel and its depth, we can invert this mapping to compute the corresponding 3D point $\mathbf{x} = \mathbf{R}^{T}(\mathbf{K}^{-1}\mathbf{u}-\mathbf{t})$.

\paragraph{Feature Extraction}
We extract 2D feature maps from the source view based on a fully convolutional feature extraction network.
The image is first downsampled by a series of stride-2 convolutions until a resolution of $64 \times 64$ is reached.
A 2D U-Net architecture \cite{RFB15a} then extracts a $64 \times 64$ feature map that is the input to the subsequent volume lifting.

\paragraph{Lifting 2D Features to 3D Observations}
The lifting layer lifts 2D features into a temporary 3D volume, representing a single 3D observation, which is then integrated into the persistent DeepVoxels representation.
We position the 3D feature volume in world space such that its center roughly aligns with the scene's center of gravity, which can be obtained cheaply from the keypoint point cloud obtained from sparse bundle adjustment.
The spatial extent is set such that the complete scene is inside the volume. 
We try to bound the scene as tightly as possible to not lose spatial resolution.
Lifting is implemented by a gathering operation.
For each voxel, the world space position of its center is projected to the source view's image space following Eq.~\ref{eq:projection}.
We extract a feature vector from the feature map using bilinear sampling and store the result in the code vector associated with the voxel.
Note, our approach is based only on a set of registered multi-view images and we do not have access to the scene geometry or depth maps, rather our approach learns automatically to resolve the depth ambiguity based on a gated recurrent network in 3D.

\paragraph{Integrating Lifted Features into DeepVoxels}
Lifted observations are integrated into the DeepVoxels representation via an integration network that is based on gated recurrent units (GRUs) \cite{Cho2014}.
In contrast to the standard application of GRUs, the integration network operates on the same volume across the full training procedure, i.e., the hidden state is \emph{persistent} across all training steps and never reset, leading to a geometrically consistent representation of the whole training corpus.
We use a uniform volumetric grid of size $w \times h \times d$ voxels, where each voxel has $f$ feature channels, i.e., the stored code vector has size $f$.
We employ one gated recurrent unit for each voxel, such that at each time step, all the features in a voxel have to be updated jointly.
The goal of the gated recurrent units is to incrementally fuse the lifted features and the hidden state during training, such that the best persistent 3D volumetric feature representation is discovered.
The gated recurrent units implement the mapping
\begin{align}
\mathbf{Z}_t &= \sigma(\mathbf{W}_z \mathbf{X}_t+ \mathbf{U}_z \mathbf{H}_{t-1}+\mathbf{B}_z) \enspace{,} \\
\mathbf{R}_t &= \sigma(\mathbf{W}_r \mathbf{X}_t+ \mathbf{U}_r \mathbf{H}_{t-1}+\mathbf{B}_r) \enspace{,}\\ 
\mathbf{S}_t &= \text{ReLU}(\mathbf{W}_s \mathbf{X}_t + \mathbf{U}_s(\mathbf{R}_t \circ \mathbf{H}_{t-1}) + \mathbf{B}_s)  \enspace{,}\\
\textbf{H}_t &= (1-\mathbf{Z}_t)\circ\mathbf{H}_{t-1} + \mathbf{Z}_t\circ\mathbf{S}_t \enspace{.}
\end{align}
Here, $\mathbf{X}_t$ is the lifted 3D feature volume of the current timestep $t$, the $\mathbf{W}_\bullet$ and $\mathbf{U}_\bullet$ are trainable 3D convolution weights, and the $\mathbf{B}_\bullet$ are trainable tensors of biases.
We follow Cho et al.~\cite{Cho2014} and employ a sigmoid activation $\sigma$ to compute the response of the tensor of update gates $\mathbf{Z}_t$ and reset gates $\mathbf{R}_t$. 
Based on the previous hidden state $\mathbf{H}_{t-1}$, the per-voxel reset values $\mathbf{R}_t$, and the lifted 3D feature volume $\mathbf{X}_t$, the tensor of new feature proposals $\mathbf{S}_t$ for the current time step $t$ is computed. $\mathbf{U}_s$ and $\mathbf{W}_s$ are single 3D convolutional layers. 
The new hidden state $\mathbf{H}_t$, the DeepVoxels representation for the current time step, is computed as a per-voxel linear combination of the old state $\mathbf{H}_{t-1}$ and the new DeepVoxel proposal $\mathbf{S}_t$.
The GRU performs one update step per lifted observation.
Afterwards, we apply a 3D inpainting U-Net that learns to fill holes in this feature representation.
At test time, only the optimally learned persistent 3D volumetric features, the DeepVoxels, are used to form the image corresponding to a novel target view. 
The 2D feature extraction, lifting layer and GRU gates are discarded and are not required for inference, see Fig.~\ref{fig:pipeline_overview}.

\paragraph{Projection Layer}
The projection layer implements the inverse of the lifting layer, i.e., it maps the 3D code vectors to the canonical coordinate system of the target view, see Fig.~\ref{fig:occlusion_net} (left).
Projection is also implemented based on a gathering operation.
For each voxel of the canonical view volume, its corresponding position in the persistent world space voxel grid is computed.
An interpolated code vector is then extracted via a trilinear interpolation and stored in the feature channels of the canonical view volume.

\paragraph{Occlusion Module}
Occlusion reasoning is essential for correct image formation and generalization to novel viewpoints. 
To this end, we propose a dedicated occlusion network that computes soft visibility for each voxel.
Each pixel in the target view is represented by one column of voxels in the canonical view volume, see Fig.~\ref{fig:occlusion_net} (left).
First, this column is concatenated with a feature column encoding the distance of each voxel to the camera, similar as in ~\cite{liu2018intriguing}. 
This allows the occlusion network to reason about voxel order.
The feature vector of each voxel in this canonical view volume is then compressed to a low-dimensional feature vector of dimension 4 by a single 3D convolutional layer.
This compressed volume is input to a 3D U-Net for occlusion reasoning.
For each ray, represented by a single-pixel column, this network predicts a scalar per-voxel visibility weight based on a softmax activation, see Fig.~\ref{fig:occlusion_net} (middle).
The canonical view volume is then flattened along the depth dimension with a weighted average, using the predicted visibility values.
The softmax weights can further be used to compute a depth map, which provides insight into the occlusion reasoning of the network, see Fig.~\ref{fig:occlusion_net} (right).

\paragraph{Rendering and Loss}
The rendering network is a mirrored version of the feature extraction network with higher capacity.
A 2D U-Net architecture takes as input the flattened canonical view volume from the occlusion network and provides reasoning across the full image, before a number of transposed convolutions directly regress the pixel values of the novel view.
We train our persistent DeepVoxels representation based on a combined $\ell_1$-loss and adversarial cross entropy loss~\cite{GoodfPMXWOCB2014}.
We found that an adversarial loss accelerates the generation of high-frequency detail earlier on in training.
Our adversarial discriminator is a fully convolutional patch-based discriminator~\cite{WangLZTKC2018}. 
We solve the resulting minimax optimization problem using ADAM~\cite{kingma2014adam}.

%% file: sections/analysis.tex
\begin{figure*}[t]
	\includegraphics[width=\linewidth]{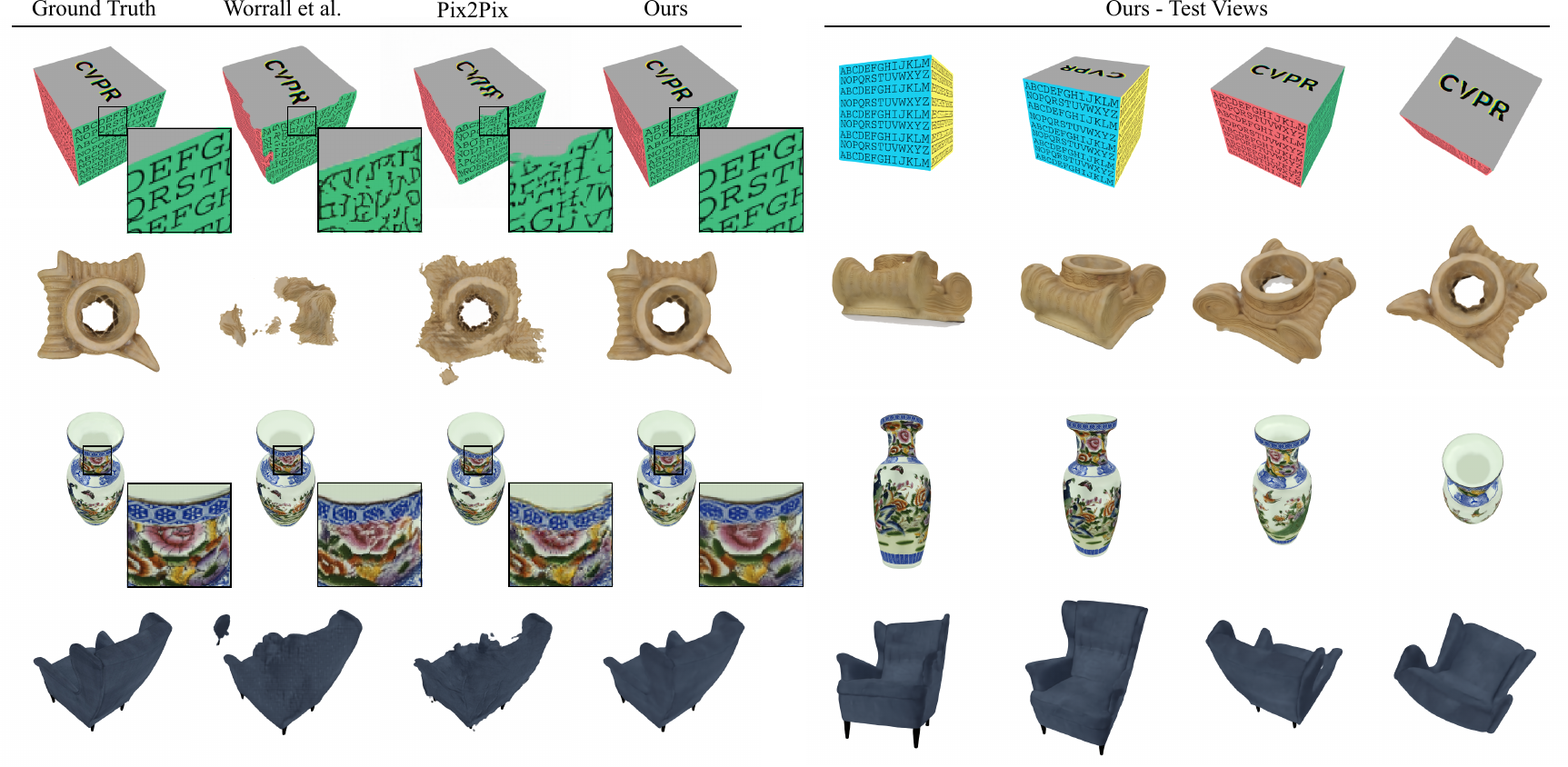}
	\vspace{-0.7cm}
	\captionof{figure}{
		Left: Comparison of the best three performing models to ground truth. From Left to right: Ground truth, Worrall et al.~\cite{worrall2017interpretable}, Isola et al.~\cite{IsolaZZE2017} (Pix2Pix), and ours. Our outputs are closest to the ground truth, performing well even in challenging cases such as the strongly foreshortened letters on the cube or the high-frequency detail of the vase. Right: Other samples of novel views generated by our model.
	}
	\label{fig:simulated_results}
\end{figure*}

\begin{table*}[t]
	\centering
	\begin{tabular}{lccccccc}
		& 
		Vase	& 
		Pedestal & 
		Chair & 
		Cube & 
		Mean \\
		&  
		PSNR / SSIM & 
		PSNR / SSIM & 
		PSNR / SSIM &
		PSNR / SSIM &
		PSNR / SSIM \\  
		\hline
		$\!\!\!\!$Nearest Neighbor	 				& $23.26$ / $0.92$  				 & $21.49$ / $0.87$  				   & $20.69$ / $0.94$  					 & $18.32$ / $0.83$ 		& $20.94$ / $0.89$ \\ 
		$\!\!\!\!$Tatarchenko et al.
		\cite{tatarchenko2015single}	 			& $22.28$ / $0.91$ 					 & $23.25$ / $0.89$ 				   & $20.22$ / $0.95$ 					 & $19.12$ / $0.84$ & $21.22$ / $0.90$\\ 	
		$\!\!\!\!$Worrall et al.~\
		\cite{worrall2017interpretable}				& $23.41$ / $0.92$  				 & $22.70$ / $0.89$    				   & $19.52$ / $0.94$    				 & $19.23$ / $0.85$ & $21.22$ / $0.90$\\ 
		$\!\!\!\!$Pix2Pix (Isola et al.)~\
		\cite{IsolaZZE2017}	& $26.36$ / $0.95$ 					 & $25.41$ / $0.91$  				   & $23.04$ / $0.96$ 					 & $19.69$ / $0.86$  & $23.63$ / $0.92$\\ 
		$\!\!\!\!$Ours			& $\mathbf{27.99}$ / $\mathbf{0.96}$ & $\mathbf{32.35}$ / $\mathbf{0.97}$  & $\mathbf{33.45}$ / $\mathbf{0.99}$  & $\mathbf{28.42}$ / $\mathbf{0.97}$ & $\mathbf{30.55}$ / $\mathbf{0.97}$ \\
	\end{tabular}
	\caption{Quantitative comparison to four baselines. Our approach obtains the best results in terms of PSNR and SSIM on all objects.}
	\label{tbl:simulated} 
	\vspace{-0.5cm}
\end{table*}

In this section, we demonstrate that DeepVoxels is a rich and semantically meaningful 3D scene representation that allows high-quality re-rendering from novel views.
First, we present qualitative and quantitative results on synthetic renderings of high-quality 3D scans of real-world objects, and compare the performance to strong machine-learning baselines with increasing reliance on geometrically structured latent spaces.
Next, we demonstrate that DeepVoxels can also be used to generate novel views on a variety of real captures, even if these scenes may violate the Lambertian assumption.
Finally, we demonstrate quantitative and qualitative benefits of explicitly reasoning about voxel visibility via the occlusion module, as well as improved model interpretability.
Please see the supplement for further studies on the sensitivity to the number of training images, the size of the voxel volume, as well as noisy camera poses.

\paragraph{Dataset and Metrics}
We evaluate model performance on synthetic data obtained from rendering $4$ high-quality 3D scans (see Fig.~\ref{fig:simulated_results}).
We center each scan at the origin and scale it to lie within the unit cube.
For the training set, we render the object from 479 poses uniformly distributed on the northern hemisphere. For the test set, we render 1000 views on an Archimedean spiral on the northern hemisphere.
All images are rendered in a resolution of $1024\times1024$ and then resized using area averaging to $512\times512$ to minimize aliasing.
We evaluate reconstruction error in terms of PSNR and SSIM~\cite{wang2004image}.

\paragraph{Implementation}
All models are implemented in PyTorch \cite{paszke2017automatic}.
Unless specified otherwise, we use a cube volume with $32^3$ voxels.
We average the $\ell_1$ loss over all pixels in the image. The $\ell_1$ and adversarial loss are weighted $200:1$.
Models are trained until convergence using ADAM with a learning rate of $4\cdot10^{-4}$. One model is trained per scene.
The proposed architecture has 170 million parameters.
At test time, rendering a single frame takes $71$ms.

\paragraph{Baselines}
We compare to three strong baselines with increasing reliance on geometry-aware latent spaces. 
The first baseline is a Pix2Pix architecture \cite{IsolaZZE2017} that receives as input the per-pixel view direction, i.e., the normalized, world-space vector from camera origin to each pixel, and is trained to translate these images into the corresponding color image.
This baseline is representative of recent achievements in 2D image-to-image translation.
The second baseline is a deep autoencoder that receives as input one of the top-$5$ nearest neighbors of the target view, and the pose of both the target and the input view are concatenated in the deep latent space, as proposed by Tatarchenko et al.~\cite{tatarchenko2015single}.
The inputs of this model at training time are thus identical to those of our model.
The third baseline learns an interpretable, rotation-equivariant latent space via the method proposed in~\cite{worrall2017interpretable, cohen2014transformation} and used previously in ~\cite{rhodin2018unsupervised}, by being fed one of the top-$5$ nearest neighbor views and then rotating the latent embedding with the rotation matrix that transforms the input to the output pose.
At test time, the previous two baselines receive the top-$1$ nearest neighbor to supply the model with the most relevant information.
We approximately match the number of parameters of each network, with all baselines having equally or slightly more parameters than our model. We train all baselines to convergence with the same loss function.
For the exact baseline architectures and number of parameters, please see the supplement.

\paragraph{Object-specific Novel View Synthesis}
We train our network and all baselines on synthetic renders of four high-quality 3D scans.
Table~\ref{tbl:simulated} compares PSNR and SSIM of the proposed architecture and the baselines.
The best-performing baseline is Pix2Pix \cite{IsolaZZE2017}.
This is surprising, since no geometrical constraints are enforced, as opposed to the approach by Worrall et al.~\cite{worrall2017interpretable}.
The proposed architecture with strongly structured latent space outperforms all baselines by a wide margin of an average $7$dB.
Fig.~\ref{fig:simulated_results} shows a qualitative comparison as well as further novel views sampled from the proposed model.
The proposed model displays robust 3D reasoning that does not break down even in challenging cases.
Notably, other models have a tendency to ``snap'' onto views seen in the training set, while the proposed model smoothly follows the test trajectory.
Please see the supplemental video for a demonstration of this behavior.
We hypothesize that this improved generalization to unseen views is due to the explicit multi-view constraints enforced by the proposed latent space.
The baseline models are not explicitly enforcing projective and epipolar geometry, which may allow them to parameterize latent spaces that are not properly representing the low-dimensional manifold of rotations. 
Although the resolution of the proposed voxel grid is $16$ times smaller than the image resolution, our model succeeds in capturing fine detail much smaller than the size of a single voxel, such as the letters on the sides of the cube or the detail on the vase. This may be due to the use of trilinear interpolation in the lifting and projection steps, which allow for a fine-grained representation to be learned.
Please see the video for full sequences, and the supplemental material for two additional synthetic scenes.

\vspace{-0.2cm}
\paragraph{Voxel Embedding vs.~Rotation-Equivariant Embedding}
As reflected in Tab.~\ref{tbl:simulated}, we outperform~\cite{worrall2017interpretable} by a wide margin both qualitatively and quantitatively.
The proposed model is constrained through multi-view geometry, while~\cite{worrall2017interpretable} has more degrees of freedom. 
Lacking occlusion reasoning, depth maps are not made explicit. 
The model may thus parameterize latent spaces that do not respect multi-view geometry.
This increases the risk of overfitting, which we observe empirically, as the baseline snaps to nearest neighbors seen during training.
While the proposed voxel embedding is memory hungry, it is very parameter efficient. 
The use of 3D convolutions means that the parameter count is independent of the voxel grid size. Giving up spatial structure means Worrell et al.~\cite{worrall2017interpretable} abandon convolutions and use fully connected layers.
However, to achieve the same latent space size of $32^3\times64$ features would necessitate more than $4.4\cdot 10^{12}$ parameters between just the fully connected layers before and after the feature transformation layer, which is infeasible.
In contrast, the proposed 3D inpainting network only has $1.7\cdot 10^7$ parameters, five orders of magnitude less.
To address memory inefficiency, the dense grid may be replaced by a sparse alternative in the future.

\begin{figure}
	\centering
	\includegraphics[width=\columnwidth]{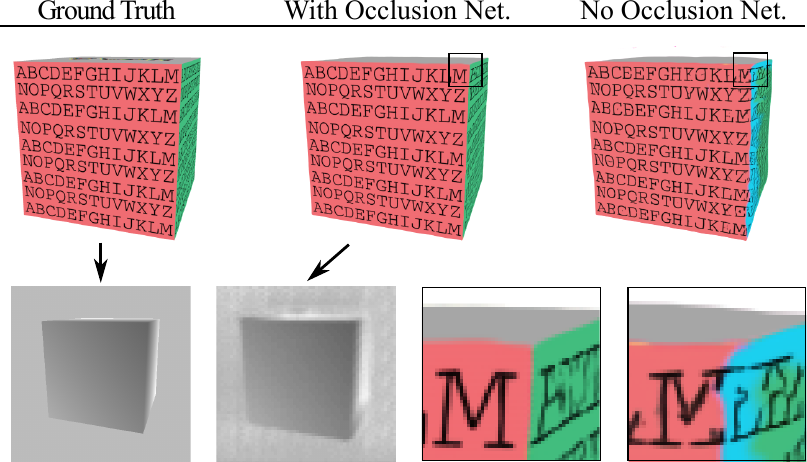}
	\caption{The occlusion module is critical to model performance. It boosts performance from $23.26$dB to $28.42$dB (cube), and from $30.02$dB to $32.35$dB (pedestal). Left: ground truth view and depth map. Center: view generated with the occlusion module and learned depth map ($64\times64$ pixels). Note that the object background is unconstrained in the depth map and may differ from ground truth. Right: without the occlusion module, the occluded, blue side of the cube (see Fig.~\ref{fig:simulated_results}) ``shines through'', and severe artifacts appear (see inset). In addition to decreasing parameter count and boosting performance, the occlusion module generates depth maps fully unsupervised, demonstrating 3D reasoning.}
	\label{fig:occlusion_net2}
	\vspace{-0.3cm}
\end{figure}

\vspace{-0.2cm}
\paragraph{Occlusion Reasoning and Interpretability} 
An essential part of the rendering pipeline is the depth test.
Similarly, the rendering network ought to be able to reason about occlusions when regressing the output view.
A naive approach might flatten the depth dimension of the canonical camera volume and subsequently reduce the number of features using a series of 2D convolutions. This leads to a drastic increase in the number of network parameters.
At training time, this further allows the network to combine features from several depths equally to regress on pixel colors in the target view.
At inference time, this results in severe artifacts and occluded parts of the object ``shining through'' (see Fig.~\ref{fig:occlusion_net2}).
Our occlusion network forces learning to use a softmax-weighted sum of voxels along each ray, which penalizes combining voxels from several depths.
As a result, novel views generated by the network with the occlusion module perform much more favorably at test time, as demonstrated in Fig.~\ref{fig:occlusion_net2}, than networks without the occlusion module.
The depth map generated by the occlusion model further demonstrates that the proposed model indeed learns the 3D structure of the scene.
We note that the depth map is learned in a fully unsupervised manner and arises out of the pure necessity of picking the most relevant voxel.
Please see the supplement for more examples of learned depth maps.
	
\begin{figure}
	\centering
	\includegraphics[width=\linewidth]{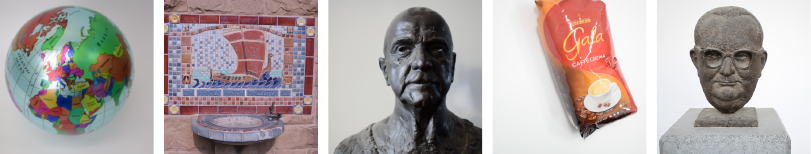}
	\caption
	{
		Novel views of real captures. Please refer to the video for full sequences with nearest neighbor comparisons.
	}
	\label{fig:real_results}
	\vspace{-0.4cm}
\end{figure}

\paragraph{Novel View Synthesis for Real Captures}
We train our network on real captures obtained with a DSLR camera. Camera poses, intrinsic camera parameters and keypoint point clouds are obtained via sparse bundle adjustment. 
The voxel grid origin is set to the respective point cloud's center of gravity. 
Voxel grid resolution is set to $64$. Each voxel stores $8$ feature channels. 
Test trajectories are obtained by linearly interpolating two randomly chosen training poses. Scenes depict a drinking fountain, two busts, a globe, and a bag of coffee. See Fig.~\ref{fig:real_results} for example model outputs. 
The drinking fountain and the globe have noticeable specularities, which are handled gracefully.
While the coffee bag is generally represented faithfully, inconsistencies appear on its highly specular surface.
Generally, results are of high quality, and only details that are significantly smaller than a single voxel, such as the tiles in the sink of the fountain, show artifacts. 
Please refer to the supplemental video for detailed results as well as a nearest-neighbor baseline.

%% file: sections/limitations.tex
Although we have demonstrated high-quality view synthesis results for a variety of challenging scenes, the proposed approach still has limitations that can be tackled in the future.
By construction, the employed 3D volume is memory inefficient, thus we have to trade local resolution for spatial extent.
The proposed model can be trained with a voxel resolution of $64^3$ with $8$ feature channels, filling a GPU with $12$GB of memory.
Future work on sparse neural networks may replace the dense representation at the core.
Please note, compelling results can already be achieved with quite small volume resolutions.
Synthesizing images from viewpoints that are significantly different from the training set, i.e., generalization, is challenging for all learning-based approaches.
While this is also true for DeepVoxels and detail is lost when viewing scenes from poses far away from training poses, DeepVoxels generally deteriorates gracefully and the 3D structure of the scene is preserved.
Please refer to the supplemental material for failure cases as well as examples of pose extrapolation.

%% file: sections/discussion.tex
We have proposed a novel 3D-structured scene representation, called DeepVoxels, that encodes the view-dependent appearance of a 3D scene using only 2D supervision.
Our approach is a first step towards 3D-structured neural scene representations and the goal of overcoming the fundamental limitations of existing 2D generative models by introducing native 3D operations into the network.

%% file: sections/acknowledgements.tex
\vspace{0.35cm}
\footnotesize
\noindent\textbf{Acknowledgements:}
We thank Robert Konrad, Nitish Padmanaban, and Ludwig Schubert for fruitful discussions, and Robert Konrad for the video voiceover.
Vincent Sitzmann was supported by a Stanford Graduate Fellowship.
Michael Zollh\"ofer and Vincent Sitzmann were supported by the Max Planck Center for Visual Computing and Communication (MPC-VCC).
Gordon Wetzstein was supported by a National Science Foundation CAREER award (IIS 1553333), by a Sloan Fellowship, and by an Okawa Research Grant.
Matthias Nie{\ss}ner and Justus Thies were supported by a Google Research Grant, the ERC Starting Grant Scan2CAD (804724), a TUM-IAS Rudolf M\"o{\ss}bauer Fellowship (Focus Group  Visual Computing), and a Google Faculty Award.

%% file: main.bbl
\begin{thebibliography}{10}\itemsep=-1pt

\bibitem{agarwal2009building}
S.~Agarwal, N.~Snavely, I.~Simon, S.~M. Seitz, and R.~Szeliski.
\newblock Building rome in a day.
\newblock In {\em Proc. CVPR}, pages 72--79, 2009.

\bibitem{Chan2018}
C.~{Chan}, S.~{Ginosar}, T.~{Zhou}, and A.~A. {Efros}.
\newblock {Everybody Dance Now}.
\newblock {\em ArXiv e-prints}, 2018.

\bibitem{chen1993view}
S.~E. Chen and L.~Williams.
\newblock View interpolation for image synthesis.
\newblock In {\em Proc. ACM SIGGRAPH}, pages 279--288, 1993.

\bibitem{Cho2014}
K.~Cho, B.~van Merrienboer, {\c{C}}.~G{\"{u}}l{\c{c}}ehre, F.~Bougares,
  H.~Schwenk, and Y.~Bengio.
\newblock Learning phrase representations using {RNN} encoder-decoder for
  statistical machine translation.
\newblock {\em CoRR}, abs/1406.1078, 2014.

\bibitem{choy20163d}
C.~B. Choy, D.~Xu, J.~Gwak, K.~Chen, and S.~Savarese.
\newblock 3d-r2n2: A unified approach for single and multi-view 3d object
  reconstruction.
\newblock In {\em Proc. ECCV}, pages 628--644, 2016.

\bibitem{cohen2014transformation}
T.~S. Cohen and M.~Welling.
\newblock Transformation properties of learned visual representations.
\newblock {\em arXiv preprint arXiv:1412.7659}, 2014.

\bibitem{dosovitskiy2017learning}
A.~Dosovitskiy, J.~T. Springenberg, M.~Tatarchenko, and T.~Brox.
\newblock Learning to generate chairs, tables and cars with convolutional
  networks.
\newblock {\em IEEE Trans. PAMI}, 39(4):692--705, 2017.

\bibitem{eigen2014depth}
D.~Eigen, C.~Puhrsch, and R.~Fergus.
\newblock Depth map prediction from a single image using a multi-scale deep
  network.
\newblock In {\em Proc. NIPS}, pages 2366--2374, 2014.

\bibitem{eslami2018neural}
S.~A. Eslami, D.~J. Rezende, F.~Besse, F.~Viola, A.~S. Morcos, M.~Garnelo,
  A.~Ruderman, A.~A. Rusu, I.~Danihelka, K.~Gregor, et~al.
\newblock Neural scene representation and rendering.
\newblock {\em Science}, 360(6394):1204--1210, 2018.

\bibitem{falorsi2018explorations}
L.~Falorsi, P.~de~Haan, T.~R. Davidson, N.~De~Cao, M.~Weiler, P.~Forr{\'e}, and
  T.~S. Cohen.
\newblock Explorations in homeomorphic variational auto-encoding.
\newblock {\em ICML Workshops}, 2018.

\bibitem{flynn2016deepstereo}
J.~Flynn, I.~Neulander, J.~Philbin, and N.~Snavely.
\newblock Deepstereo: Learning to predict new views from the world's imagery.
\newblock In {\em Proc. CVPR}, pages 5515--5524, 2016.

\bibitem{furukawa2010accurate}
Y.~Furukawa and J.~Ponce.
\newblock Accurate, dense, and robust multiview stereopsis.
\newblock {\em IEEE Trans. PAMI}, 32(8):1362--1376, 2010.

\bibitem{GoodfPMXWOCB2014}
I.~J. Goodfellow, J.~Pouget-Abadie, M.~Mirza, B.~Xu, D.~Warde-Farley, S.~Ozair,
  A.~Courville, and Y.~Bengio.
\newblock Generative adversarial nets.
\newblock In {\em Proc. NIPS}, 2014.

\bibitem{greene1986environment}
N.~Greene.
\newblock Environment mapping and other applications of world projections.
\newblock {\em IEEE CG\&A}, 6(11):21--29, 1986.

\bibitem{Hartley:2003}
R.~Hartley and A.~Zisserman.
\newblock {\em Multiple View Geometry in Computer Vision}.
\newblock Cambridge University Press, 2nd edition, 2003.

\bibitem{hedman2018deepblending}
P.~Hedman, J.~Philip, T.~Price, J.-M. Frahm, G.~Drettakis, and G.~Brostow.
\newblock Deep blending for free-viewpoint image-based rendering.
\newblock {\em ACM Trans. Graph. (SIGGRAPH Asia)}, 37(6), 2018.

\bibitem{hedman2016scalable}
P.~Hedman, T.~Ritschel, G.~Drettakis, and G.~Brostow.
\newblock Scalable inside-out image-based rendering.
\newblock {\em ACM Trans. Graph. (SIGGRAPH Asia)}, 35(6):231, 2016.

\bibitem{henriques2018mapnet}
J.~F. Henriques and A.~Vedaldi.
\newblock Mapnet: An allocentric spatial memory for mapping environments.
\newblock In {\em Proc. CVPR}, pages 8476--8484, 2018.

\bibitem{HintoS2006}
G.~E. Hinton and R.~Salakhutdinov.
\newblock Reducing the dimensionality of data with neural networks.
\newblock {\em Science}, 313(5786):504--507, July 2006.

\bibitem{IsolaZZE2017}
P.~Isola, J.-Y. Zhu, T.~Zhou, and A.~A. Efros.
\newblock Image-to-image translation with conditional adversarial networks.
\newblock In {\em Proc. CVPR}, pages 5967--5976, 2017.

\bibitem{Jack2018}
D.~Jack, J.~K. Pontes, S.~Sridharan, C.~Fookes, S.~Shirazi, F.~Maire, and
  A.~Eriksson.
\newblock Learning free-form deformations for 3d object reconstruction.
\newblock {\em CoRR}, abs/1803.10932, 2018.

\bibitem{Jaderberg2015}
M.~Jaderberg, K.~Simonyan, A.~Zisserman, and k.~kavukcuoglu.
\newblock Spatial transformer networks.
\newblock In {\em Proc. NIPS}, pages 2017--2025. 2015.

\bibitem{NIPS2016_6600}
D.~Jimenez~Rezende, S.~M.~A. Eslami, S.~Mohamed, P.~Battaglia, M.~Jaderberg,
  and N.~Heess.
\newblock Unsupervised learning of 3d structure from images.
\newblock In {\em Proc. NIPS}, pages 4996--5004. 2016.

\bibitem{kalantari2016learning}
N.~K. Kalantari, T.-C. Wang, and R.~Ramamoorthi.
\newblock Learning-based view synthesis for light field cameras.
\newblock {\em ACM Trans. Graph. (SIGGRAPH Asia)}, 35(6):193, 2016.

\bibitem{kar2017learning}
A.~Kar, C.~H{\"a}ne, and J.~Malik.
\newblock Learning a multi-view stereo machine.
\newblock In {\em Proc. NIPS}, pages 365--376, 2017.

\bibitem{KarraALL2018}
T.~Karras, T.~Aila, S.~Laine, and J.~Lehtinen.
\newblock Progressive growing of {GANs} for improved quality, stability, and
  variation.
\newblock In {\em Proc. ICLR}, 2018.

\bibitem{Kazhdan:2006}
M.~Kazhdan, M.~Bolitho, and H.~Hoppe.
\newblock Poisson surface reconstruction.
\newblock In {\em Proc. SGP}, pages 61--70, 2006.

\bibitem{kim2018DeepVideo}
H.~Kim, P.~Garrido, A.~Tewari, W.~Xu, J.~Thies, N.~Nie{\ss}ner, P.~P{\'e}rez,
  C.~Richardt, M.~Zollh{\"o}fer, and C.~Theobalt.
\newblock {Deep Video Portraits}.
\newblock {\em ACM Trans. Graph. (SIGGRAPH)}, 2018.

\bibitem{kingma2014adam}
D.~P. Kingma and J.~Ba.
\newblock Adam: A method for stochastic optimization.
\newblock {\em arXiv preprint arXiv:1412.6980}, 2014.

\bibitem{jKingma2014}
D.~P. Kingma and M.~Welling.
\newblock Auto-encoding variational bayes.
\newblock {\em CoRR}, abs/1312.6114, 2013.

\bibitem{Kulkarni2015}
T.~D. Kulkarni, W.~F. Whitney, P.~Kohli, and J.~Tenenbaum.
\newblock Deep convolutional inverse graphics network.
\newblock In {\em Proc. NIPS}, pages 2539--2547. 2015.

\bibitem{lin2018learning}
C.-H. Lin, C.~Kong, and S.~Lucey.
\newblock Learning efficient point cloud generation for dense 3d object
  reconstruction.
\newblock In {\em AAAI}, 2018.

\bibitem{lippman1980movie}
A.~Lippman.
\newblock Movie-maps: An application of the optical videodisc to computer
  graphics.
\newblock In {\em ACM SIGGRAPH}, volume~14, pages 32--42, 1980.

\bibitem{liu2018intriguing}
R.~Liu, J.~Lehman, P.~Molino, F.~P. Such, E.~Frank, A.~Sergeev, and
  J.~Yosinski.
\newblock An intriguing failing of convolutional neural networks and the
  coordconv solution.
\newblock {\em arXiv preprint arXiv:1807.03247}, 2018.

\bibitem{Maturana2015}
D.~Maturana and S.~Scherer.
\newblock Voxnet: A 3d convolutional neural network for real-time object
  recognition.
\newblock In {\em Proc. IROS}, page 922 – 928, September 2015.

\bibitem{MirzaO2014}
M.~Mirza and S.~Osindero.
\newblock Conditional generative adversarial nets.
\newblock arXiv:1411.1784, 2014.

\bibitem{RenderNet}
T.~H. Nguyen-Phuoc, C.~Li, S.~Balaban, and Y.~Yang.
\newblock Rendernet: A deep convolutional network for differentiable rendering
  from 3d shapes.
\newblock In {\em Proc. NIPS 2018}, pages 7902--7912. 2018.

\bibitem{Oord:2016}
A.~v.~d. Oord, N.~Kalchbrenner, O.~Vinyals, L.~Espeholt, A.~Graves, and
  K.~Kavukcuoglu.
\newblock Conditional image generation with pixelcnn decoders.
\newblock In {\em Proc. NIPS}, pages 4797--4805, 2016.

\bibitem{Park17}
E.~Park, J.~Yang, E.~Yumer, D.~Ceylan, and A.~C. Berg.
\newblock Transformation-grounded image generation network for novel 3d view
  synthesis.
\newblock {\em CoRR}, abs/1703.02921, 2017.

\bibitem{paszke2017automatic}
A.~Paszke, S.~Gross, S.~Chintala, G.~Chanan, E.~Yang, Z.~DeVito, Z.~Lin,
  A.~Desmaison, L.~Antiga, and A.~Lerer.
\newblock Automatic differentiation in pytorch.
\newblock In {\em NIPS-W}, 2017.

\bibitem{penner2017soft}
E.~Penner and L.~Zhang.
\newblock Soft 3d reconstruction for view synthesis.
\newblock {\em ACM Trans. Graph. (SIGGRAPH Asia)}, 36(6):235, 2017.

\bibitem{qi2017pointnet}
C.~R. Qi, H.~Su, K.~Mo, and L.~J. Guibas.
\newblock Pointnet: Deep learning on point sets for 3d classification and
  segmentation.
\newblock {\em Proc. CVPR}, 2017.

\bibitem{qi2016volumetric}
C.~R. Qi, H.~Su, M.~Nie{\ss}ner, A.~Dai, M.~Yan, and L.~Guibas.
\newblock Volumetric and multi-view cnns for object classification on 3d data.
\newblock In {\em Proc. CVPR}, 2016.

\bibitem{RadfoMC2016}
A.~Radford, L.~Metz, and S.~Chintala.
\newblock Unsupervised representation learning with deep convolutional
  generative adversarial networks.
\newblock In {\em Proc. ICLR}, 2016.

\bibitem{rhodin2018unsupervised}
H.~Rhodin, M.~Salzmann, and P.~Fua.
\newblock Unsupervised geometry-aware representation for 3d human pose
  estimation.
\newblock {\em Proc. ECCV}, 2018.

\bibitem{Riegler2017OctNet}
G.~Riegler, A.~O. Ulusoy, and A.~Geiger.
\newblock Octnet: Learning deep 3d representations at high resolutions.
\newblock In {\em Proc. CVPR}, 2017.

\bibitem{RonneFB2015}
O.~Ronneberger, P.~Fischer, and T.~Brox.
\newblock {U-Net}: Convolutional networks for biomedical image segmentation.
\newblock In {\em Proc. MICCAI}, pages 234--241, 2015.

\bibitem{RFB15a}
O.~Ronneberger, P.Fischer, and T.~Brox.
\newblock U-net: Convolutional networks for biomedical image segmentation.
\newblock In {\em Proc. MICCAI}, pages 234--241, 2015.

\bibitem{schoenberger2016sfm}
J.~L. Sch\"{o}nberger and J.-M. Frahm.
\newblock Structure-from-motion revisited.
\newblock In {\em Proc. CVPR}, 2016.

\bibitem{schoenberger2016mvs}
J.~L. Sch\"{o}nberger, E.~Zheng, M.~Pollefeys, and J.-M. Frahm.
\newblock Pixelwise view selection for unstructured multi-view stereo.
\newblock In {\em Proc. ECCV}, 2016.

\bibitem{shum2000review}
H.~Shum and S.~B. Kang.
\newblock Review of image-based rendering techniques.
\newblock In {\em Proc. VCIP}, pages 2--14, 2000.

\bibitem{snavely2006photo}
N.~Snavely, S.~M. Seitz, and R.~Szeliski.
\newblock Photo tourism: exploring photo collections in 3d.
\newblock In {\em ACM Trans. Graph. (SIGGRAPH)}, volume~25, pages 835--846,
  2006.

\bibitem{szeliski2010computer}
R.~Szeliski.
\newblock {\em Computer vision: algorithms and applications}.
\newblock Springer Science \& Business Media, 2010.

\bibitem{tatarchenko2015single}
M.~Tatarchenko, A.~Dosovitskiy, and T.~Brox.
\newblock Single-view to multi-view: Reconstructing unseen views with a
  convolutional network.
\newblock {\em CoRR abs/1511.06702}, 1(2):2, 2015.

\bibitem{Triggs:1999}
B.~Triggs, P.~F. McLauchlan, R.~I. Hartley, and A.~W. Fitzgibbon.
\newblock Bundle adjustment - a modern synthesis.
\newblock In {\em Proc. ICCV Workshops}, pages 298--372, 2000.

\bibitem{lsiTulsiani18}
S.~Tulsiani, R.~Tucker, and N.~Snavely.
\newblock Layer-structured 3d scene inference via view synthesis.
\newblock In {\em Proc. ECCV}, 2018.

\bibitem{drcTulsiani17}
S.~Tulsiani, T.~Zhou, A.~A. Efros, and J.~Malik.
\newblock Multi-view supervision for single-view reconstruction via
  differentiable ray consistency.
\newblock In {\em Proc. CVPR}, 2017.

\bibitem{WangLZTKC2018}
T.-C. Wang, M.-Y. Liu, J.-Y. Zhu, A.~Tao, J.~Kautz, and B.~Catanzaro.
\newblock High-resolution image synthesis and semantic manipulation with
  conditional {GANs}.
\newblock In {\em Proc. CVPR}, 2018.

\bibitem{wang2004image}
Z.~Wang, A.~C. Bovik, H.~R. Sheikh, and E.~P. Simoncelli.
\newblock Image quality assessment: from error visibility to structural
  similarity.
\newblock {\em IEEE Trans. Im. Proc.}, 13(4):600--612, 2004.

\bibitem{worrall2017interpretable}
D.~E. Worrall, S.~J. Garbin, D.~Turmukhambetov, and G.~J. Brostow.
\newblock Interpretable transformations with encoder-decoder networks.
\newblock In {\em Proc. ICCV}, volume~4, 2017.

\bibitem{yan2016perspective}
X.~Yan, J.~Yang, E.~Yumer, Y.~Guo, and H.~Lee.
\newblock Perspective transformer nets: Learning single-view 3d object
  reconstruction without 3d supervision.
\newblock In {\em Proc. NIPS}, pages 1696--1704, 2016.

\bibitem{yang2015weakly}
J.~Yang, S.~E. Reed, M.-H. Yang, and H.~Lee.
\newblock Weakly-supervised disentangling with recurrent transformations for 3d
  view synthesis.
\newblock In {\em Proc. NIPS}, pages 1099--1107, 2015.

\bibitem{zhou2016view}
T.~Zhou, S.~Tulsiani, W.~Sun, J.~Malik, and A.~A. Efros.
\newblock View synthesis by appearance flow.
\newblock In {\em Proc. ECCV}, pages 286--301. Springer, 2016.

\bibitem{Zhu2018}
H.~Zhu, H.~Su, P.~Wang, X.~Cao, and R.~Yang.
\newblock View extrapolation of human body from a single image.
\newblock {\em CoRR}, abs/1804.04213, 2018.

\end{thebibliography}
